**Enhancing Maritime Domain Awareness on Inland Waterways: A YOLO-Based Fusion of Satellite and AIS for Vessel Characterization**

**Geoffery Agorku***
Department of Civil Engineering
University of Arkansas, Fayetteville, Arkansas, 72701
gagorku@uark.edu
https://orcid.org/0009-0000-7716-0334

**Sarah Hernandez, PhD, P.E**
Department of Civil Engineering
University of Arkansas, Fayetteville, Arkansas, 72701
sarahvh@uark.edu
https://orcid.org/0000-0002-4243-1461

**Hayley Hames**
Center for Advanced Spatial Technologies
University of Arkansas, Fayetteville, Arkansas, 72701
hhames@uark.edu
https://orcid.org/0009-0002-7285-3005

**Cade Wagner**
Department of Mathematical Sciences
University of Arkansas, Fayetteville, Arkansas, 72701
cadew@uark.edu
https://orcid.org/0009-0000-1650-6450

Word Count: 6,496 words + 4 tables (250 words per table) = 7,496 words

Submitted *[10/13/2025]*
*Corresponding Author

**ABSTRACT**

Maritime Domain Awareness (MDA) for inland waterways remains challenged by cooperative system vulnerabilities. This paper presents a novel framework that fuses high-resolution satellite imagery with vessel trajectory data from the Automatic Identification System (AIS). This work addresses the limitations of AIS-based monitoring by leveraging non-cooperative satellite imagery and implementing a fusion approach that links visual detections with AIS data to identify dark vessels, validate cooperative traffic, and support advanced MDA. The You Only Look Once (YOLO) v11 object detection model is used to detect and characterize vessels and barges by vessel type, barge cover, operational status, barge count, and direction of travel. An annotated data set of 4,550 instances was developed from 5,973 mi² of Lower Mississippi River imagery. Evaluation on a held-out test set demonstrated *vessel classification* (tugboat, crane barge, bulk carrier, cargo ship, and hopper barge) with an F1-score of 95.8%; *barge cover* (covered or uncovered) detection yielded an F1-score of 91.6%; *operational status* (staged or in motion) classification reached an F1-score of 99.4%. *Directionality* (upstream, downstream) yielded 93.8% accuracy. The *barge count estimation* resulted in a mean absolute error (MAE) of 2.4 barges. Spatial transferability analysis across geographically disjoint river segments showed accuracy was maintained as high as 98%. These results underscore the viability of integrating non-cooperative satellite sensing with AIS fusion. This approach enables near-real-time fleet inventories, supports anomaly detection, and generates high-quality data for inland waterway surveillance. Future work will expand annotated datasets, incorporate temporal tracking, and explore multi-modal deep learning to further enhance operational scalability.

**Keywords:** Maritime Domain Awareness (MDA), Satellite Imagery and AIS Fusion, (You Only Look Once) YOLO Object Detection, Automatic Identification System (AIS), Inland Waterway Monitoring (IWT)

1. **INTRODUCTION**

Maritime Domain Awareness (MDA) is a pillar of global security, international commerce, and environmental stewardship (*1*). The ability to effectively monitor waterways is an operational imperative. It supports a vast array of functions, from enforcing fishing regulations and ensuring maritime traffic safety to interdicting illicit ship-to-ship transfers, monitoring foreign military movements, and safeguarding the integrity of global supply chains (*2*).

For decades, MDA has been a collection of cooperative, self-reporting systems designed to enhance navigational safety and facilitate regulatory oversight. Paramount among these cooperative systems are the Automatic Identification System (AIS) and the Vessel Monitoring System (VMS) (*1*). AIS has become a universal tool, mandated for international voyaging ships over a certain tonnage, passenger ships, and other specific vessel classes (*3*). By broadcasting dynamic information such as a vessel's unique identity, position, course, and speed, AIS forms the basis of modern vessel traffic services and collision avoidance systems (*4*). This system, along with VMS, offers a detailed, real-time view of cooperative maritime traffic, enabling a significant degree of situational awareness (*5*).

The AIS strengths are evident in a wide range of operational contexts. AIS enables real-time vessel tracking, supports collision avoidance, and global maritime traffic analysis, port planning, and regulatory oversight. While AIS has become an indispensable tool for maritime situational awareness, its limitations are increasingly targeted by illegal operators (*6*). Its effectiveness is compromised by some vulnerabilities which are not merely passive technical limitations, but active points of exploitation. The documented practice of vessels intentionally deactivating their AIS transponders, a tactic known as "going dark", is frequently employed to conceal illegal, unreported, and unregulated (IUU) fishing, smuggling, and other concealed maritime activities (*7–9*). Although such actions render the vessel temporarily invisible to cooperative systems, AIS remains an invaluable source of structured, high-frequency data for most of the compliant traffic. As such, it serves as a critical input for hybrid surveillance systems, especially when augmented with non-cooperative sensing technologies that can identify, verify, and compensate for AIS signal gaps.

Beyond deliberate evasion, AIS performance degrades under certain conditions. In high-traffic density areas such as congested ports, narrow straits, and inland waterways, the system suffers from significant signal interference, packet collisions, and data loss (*10*). This phenomenon creates a paradoxical situation where surveillance reliability decreases as vessel density and risk increase. The complex terrain and infrastructure along inland rivers further worsen these signal propagation issues, making consistent AIS tracking in these environments particularly challenging (*11*). Furthermore, the system is susceptible to data spoofing, where vessels can transmit false identity or location information to actively deceive authorities, and the raw data streams are often plagued by noise and corruption that require extensive filtering and validation to be useful (*1*).

Arguably, the most critical systemic constraint of AIS lies in its limited scope. A vast portion of the global maritime fleet, comprising primarily small-scale vessels under 15-20 meters in length, are not required to carry AIS transponders at all (*12*). This regulatory exemption creates a persistent blind spot, obscuring the activities of a large segment of vessels, namely fishing fleets and local traffic.

The limitations of AIS have created a clear operational need for an independent, non-cooperative verification layer capable of observing all vessels, regardless of their size, location, or willingness to self-report. Consequently, the development and integration of non-cooperative surveillance technologies such as high-resolution electro-optical and synthetic aperture radar (SAR) satellite imagery, represents a necessary and direct counter-approach. Satellite remote sensing provides a synoptic view over vast geographic areas with a regularity that is logistically infeasible and cost-prohibitive for traditional air or sea patrols (*13*). This technology serves as the essential "eyes in the sky," capable of detecting vessels that are either unable or unwilling to announce their presence (*14*). Satellite remote sensing is a viable and adept tool to build a more complete and resilient MDA system (*1*).

The objectives of this paper are as follows:

i. Develop a robust computer vision framework for detecting, classifying, and analyzing inland waterway vessels using high-resolution satellite imagery, with capabilities extending beyond basic object detection to include barge count estimation, cover type classification, operational status inference, and direction of movement.
ii. Address the limitations of AIS-based monitoring by leveraging non-cooperative satellite imagery and implementing a fusion approach that links visual detections with AIS data to identify dark vessels, validate cooperative traffic, and support advanced MDA.
iii. Evaluate model generalizability and scalability through hyperparameter tuning, transfer learning, and spatial transferability analysis to ensure reliable performance across geographically distinct inland river segments and support future deployment in operational monitoring systems.

## 2. BACKGROUND

This paper applies deep learning to vessel object detection using satellite imagery with the goal of fully characterizing vessels along inland waterways. The advancements in deep learning architectures for vessel detection are linked to the availability of large-scale, high-quality training datasets (*15*). The performance, robustness, and generalizability of any model is fundamentally constrained by the data upon which it is trained (*16*). Publicly available datasets for vessel detection are dominated by data collected in open-ocean or large coastal port environments (*17*). Thus, research in this context is biased towards open-ocean ("blue water") and major coastal port surveillance (*18*). This is likely due to the initial drivers for this technology being defense, national security, and large-scale commercial shipping monitoring (*19*). The recent introduction of the Multi-environment Inland Waterway Vessel Dataset (MEIWVD) represents a groundbreaking effort to directly address this gap (*20*). With its 32,478 images captured in the complex environments of the Yangtze River Basin, MEIWVD provides the first large-scale, public benchmark for this domain. Its explicit focus on capturing diverse environmental conditions, especially fog and rain, and common inland vessel types, makes it a representative testbed for models intended for inland waterway applications compared to most existing open-ocean datasets (*20*).

Consequently, there is a lack of publicly available, large-scale datasets for the unique challenges of inland waterways ("brown water") (*20*). Inland waterways are characterized by narrow, constrained channels, which means vessels are almost always seen against a backdrop of background clutter from shorelines, vegetation, bridges, buildings, and other infrastructure (*20*). The vessel types are often different, with a prevalence of barges, tugs, and smaller passenger ferries. Furthermore, these environments often have poor AIS signal quality, making non-cooperative surveillance critical (*11*). A clear gap exists for the development of robust, computationally efficient, and clutter-resistant object detection models specifically optimized for inland waterway surveillance (*20*). This research area requires a focus on creating architectures that can effectively distinguish target vessels (such as barges and tugs) from complex and proximate land-based clutter and can perform accurately despite the challenging imaging conditions (e.g., fog, low light) common to these environments.

While deep learning models have achieved up to 98% precision in detecting vessels from satellite imagery alone (*21–23*), the next limitation in enhancing MDA lies in the multi-modal fusion of non-cooperative satellite data with cooperative data streams, primarily the AIS (*24*). This synergistic approach promises to create complete, accurate, and contextually rich data than either data source can provide in isolation (*25*). Satellite imagery provides the definitive "what" and "where" for all visible vessels, while AIS provides the "who" (name, MMSI), "why" (vessel type, destination), and "how" (course, speed) for the subset of cooperative vessels (*10*). The primary application of this fusion is the identification of "dark" vessels (*26*). By correlating every vessel detected in a satellite image with a known AIS track, any remaining, uncorrelated detections can be flagged as potentially non-cooperative or "dark" vessels, which are of high interest for security and regulatory agencies (*10*).

Fusion extends beyond tracking "dark vessels". The process enables a host of high-value downstream applications. For instance, by automatically correlating image detections with the metadata in AIS messages (vessel type, length, beam), fusion can be used to generate massive, high-quality, automatically annotated training datasets for the next generation of vessel detection and classification models, bypassing the prohibitively expensive process of manual labeling (*27*). Furthermore, fusion provides the ground truth needed to validate the behavior of cooperative vessels and to detect anomalies, such as a vessel deviating from its declared course or operating in a restricted area (*28*). Improving the accuracy of data fusion therefore has a cascading positive effect across the entire ecosystem of maritime analytics (*29*).

Furthermore, a gap exists between the well-established problem of static vessel detection and the emerging operational need for dynamic, multi-object tracking of vessel fleets in time-series satellite imagery (*12*). This requires novel frameworks to effectively integrate state-of-the-art detectors with tracking algorithms (*30*). Such frameworks must handle the challenges of low-frequency observations (i.e., long gaps between images), vessel occlusions, and the problem of re-identifying the same vessel across multiple satellite passes (*31*).

Finally, the body of research in maritime surveillance is weighted towards answering the fundamental questions of "where is the vessel?" (detection) and "what kind of vessel is it?" (classification) (*32*). A higher-level, more economically valuable line of inquiry revolves around the question: "What is the vessel doing?" This involves inferring the operational status and activity of vessels and their interaction with maritime infrastructure (*33*). A research gap exists in developing methodologies that leverage vessel detection as a foundational layer to infer higher-level, actionable intelligence about operational status and economic activity (*34*). This includes research into the dynamic inventorying of waterway assets (e.g., counting active vs. moored barges), near-real-time monitoring of port throughput, and analyzing the complex interactions between vessels and fixed infrastructure. Successfully bridging this gap would transform satellite surveillance from a security-focused tool into an instrument for creating a dynamic digital twin of the global maritime transportation system.

## 3.    METHODS

Following a description of the data acquisition approach (Section 3.1), the methodology is divided into and discussed in two distinct parts: (A) developing a vessel and infrastructure detection and characterization model using computer vision (Section 3.2), (B) performing spatiotemporal data fusion of satellite and vessel trajectory data (Section 3.3) (**Figure 1**).

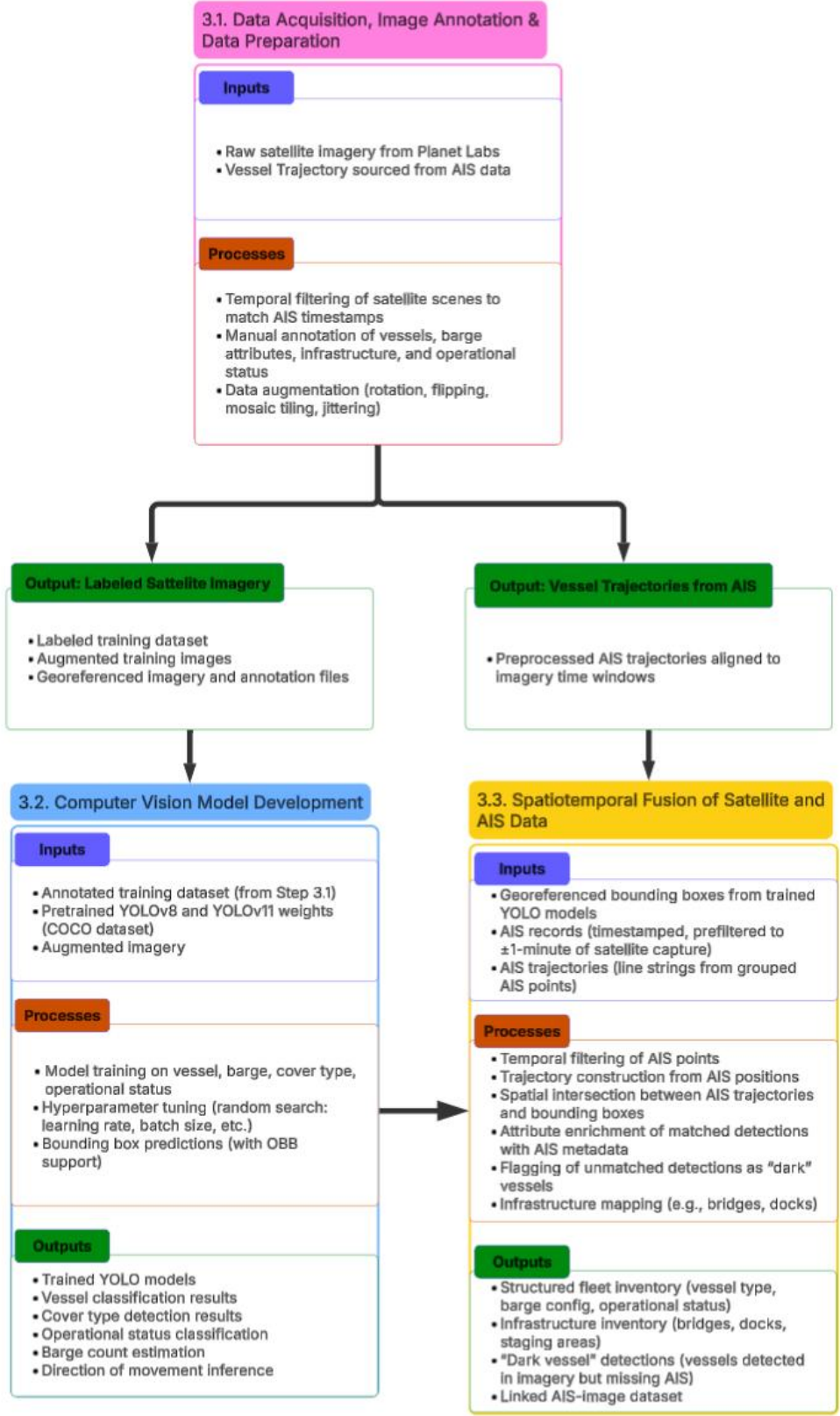

**Figure 1: Overview of the satellite–AIS fusion methodology for inland waterway vessel detection**

## 3.1. Data Acquisition, Image Annotation, and Data Preparation

Two complimentary datasets are required for implementing the methodology presented in this paper:

- *Vessel trajectory*: sources from AIS which provides vessel identities, positions, speeds, and headings; can be retrieved from public and commercial sources.
- *Satellite imagery*: sourced from commercial providers (e.g., Planet Labs, Maxar) or public platforms (e.g., Sentinel-2). To ensure temporal alignment, satellite scenes are filtered to coincide with periods of AIS coverage in the same region.

To train supervised object detection models, a representative set of satellite images is annotated with bounding boxes and class labels. Labeled object classes are:

- *Vessel types*: tug/towboats, cargo ships, and bulk carriers
- *Barge types*: hopper and crane barge
- *Barge cover*: covered and uncovered
- *Operational status*: in-motion, staged (barges positioned along the riverbank with a tug or towboat still attached), and moored/berthed (barges positioned along the riverbank or at a dock without a tug or towboat attached.)
- *Infrastructure object:* docks, bridges, and staging areas (located in the waterway)
- *Direction of movement*: upstream, downstream, and stationary

Data augmentation techniques such as image rotation, horizontal flipping, color jittering, and mosaic tiling are applied to simulate environmental variation and increase model generalizability.

## 3.2. Computer Vision Model Development

This framework employs deep learning-based object detection to identify vessels and infrastructure from satellite imagery. The You Only Look Once (YOLO) architectures YOLOv11 is selected. This model balances detection speed, accuracy, and adaptability to high-resolution remote sensing contexts. Importantly, this model supports Object Oriented Bounding Boxes (OBB) to better localize elongated vessels, particularly those aligned at arbitrary angles along winding river paths (**Figure 2**). This model is well-suited to the inland waterway domain, where objects vary widely in size and orientation and are often embedded in visually complex environments.

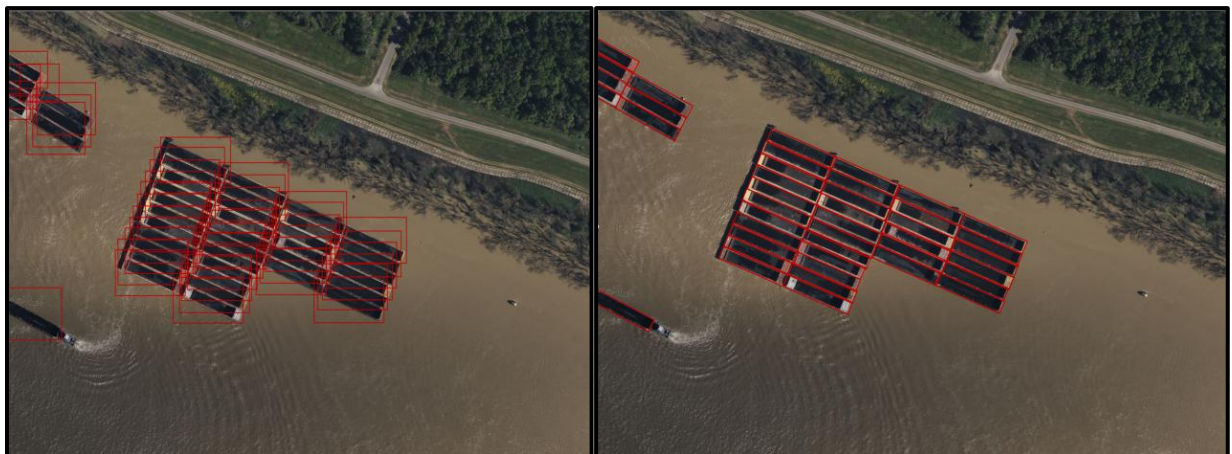

a) Normal bounding boxes          b) Object-oriented bounding boxes

**Figure 2: Bounding box types**

To accelerate training and enhance generalization, both models were initialized using transfer learning. Pretrained weights from the Common Objects in Context (COCO) dataset are used as a starting point to leverage broad visual feature representations. Fine-tuning is performed using our training dataset, allowing the models to adapt to the unique visual patterns of riverine traffic, such as elongated hulls, grouped barge configurations, and shoreline clutter.

Training is conducted using PyTorch-based YOLO implementations on NVIDIA GPUs. Model performance is evaluated across a range of subtasks relevant to inland waterway monitoring: vessel and barge classification, cover type detection, operational status classification, barge count estimation, object-to-AIS linkage accuracy, and direction-of-movement inference. Evaluation metrics include precision, recall, F1-score, accuracy, mean absolute error (MAE), and Symmetric Mean Absolute Percentage Error (sMAPE) defined as:

$$Precision = \frac{True\ Positives}{True\ Positives + False\ Positives} \quad \textbf{Equation 1}$$

$$Recall = \frac{True\ Positives}{True\ Positivese + False\ Negatives} \quad \textbf{Equation 2}$$

$$F1 - Score = \frac{True\ Positives}{True\ Positivese + False\ Negatives} \quad \textbf{Equation 3}$$

$$Accuracy = \frac{Correct\ Predictions}{All\ Predictions} \quad \textbf{Equation 4}$$

$$MAE = \frac{1}{n}\sum_{i=1}^{n}|x_i - \hat{x}_i| \quad \textbf{Equation 5}$$

$$sMAPE = \frac{100\%}{n}\sum_{i=1}^{n}\frac{2|x_i - \hat{x}_i|}{|x_i| + |\hat{x}_i|} \quad \textbf{Equation 6}$$

where

$n$ = The total number of vessel instances (e.g., tugs or towboats) in the test dataset for which barge counts are estimated.
$x_i$ = The true number of barges attached to the i-th vessel, based on manual annotation or validated ground truth.
$\hat{x}_i$ = The predicted number of barges attached to the i-th vessel, as estimated by the computer vision model.
$|x_i - x|$ = The absolute error in barge count estimation for the i-th vessel. i.e., how far off the prediction was from the actual count.

Precision (**Equation 1**) and recall (**Equation 2**) are used to evaluate detection reliability and completeness while the F1-Score (**Equation 3**) provided a balanced indicator of model performance.

Hyperparameters such as learning rate, batch size, confidence threshold, and input image size are tuned using random or grid search strategies. Learning rate scheduling and early stopping are employed to avoid overfitting.

### 3.3. Spatiotemporal Fusion of Satellite and AIS Data

To associate visual detections in the satellite imagery with AIS-detected ("cooperative") vessels, a spatiotemporal fusion process is implemented:

1. **Temporal filtering:** AIS records are selected based on their timestamp proximity to each satellite image. Once vessel detections are generated by the YOLO model and bounding boxes are georeferenced within each satellite image, we enrich satellite vessel detections with AIS information by exploiting both temporal proximity and spatial intersection. For each satellite scene, the precise acquisition timestamp is recorded and used to filter the AIS database, extracting messages whose timestamp falls within a two-minute window before and after the image capture time. This narrow temporal window accommodates minor clock discrepancies between satellite and AIS sources while minimizing false matches.
2. **Trajectory construction:** The next step reconstructs vessel trajectories from the AIS points. All AIS records are grouped by their unique vessel (e.g., MMSI) identifier and chronologically ordered by their timestamp. By sequentially connecting these points, we form continuous line strings that represent the path each vessel traversed through the imaged region. This "points-to-path" algorithm effectively transforms a cloud of time-stamped AIS coordinates into coherent trajectories, capturing both the vessel's course and speed variations over time.
3. **Spatial intersection:** To establish a link between a detected bounding box and an AIS trajectory, we perform a spatial intersection test. Each bounding box is translated into a geospatial polygon, leveraging the image's ground control information to align pixel coordinates with real-world positions. When a bounding polygon intersects one of the AIS-derived trajectories, and its timestamp lies within the same ±2-minute interval, the detection is deemed to correspond to that vessel (**Figure 3**). This dual constraint of spatial overlap and temporal alignment ensures high confidence in the match and reduces false associations, particularly in congested or multi-vessel scenes.

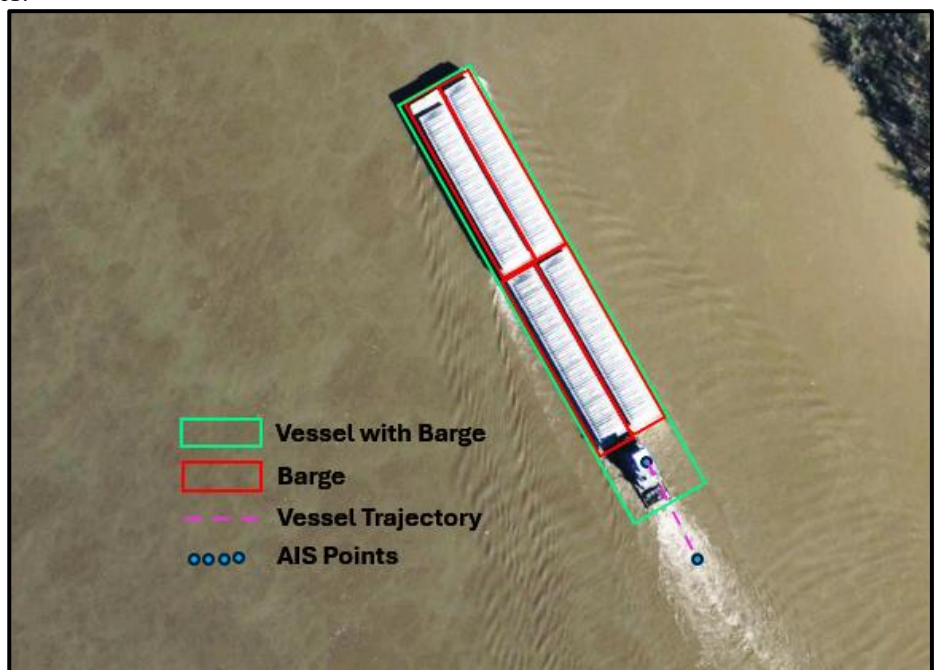

Figure 3: Example of spatiotemporal fusion of satellite imagery and AIS data

Successful matches inherit AIS metadata (e.g., vessel ID, course, speed) and are aggregated into an inventory of inland waterway assets. This inventory will include inferred characteristics from the YOLO models applied in this paper such as barge counts and cover type, operational status (e.g. moored, staged, in motion), and direction of movement. This output provides a snapshot of fleet composition, deployment patterns, and river infrastructure utilization at the time of observation.

Detections without corresponding AIS matches are flagged as potentially non-cooperative or "dark" vessels. These may include vessels that have disabled their transponders, operate without AIS mandates, or are engaged in illicit activity. This classification supports applications in anomaly detection, regulatory compliance, and MDA, particularly in regions with limited cooperative tracking coverage.

Additionally, the model is also tasked with cataloging river infrastructure features, such as bridges, barge staging areas, and docking facilities, directly from the imagery. Each detected infrastructure element is georeferenced using the same ground control data that aligns vessel bounding boxes to real-world coordinates. These features are then recorded in a dedicated infrastructure inventory, which contains their precise location information and type. This complementary dataset provides crucial context for understanding fleet operations, enabling analyses of how vessel movements relate to nearby infrastructure.

## 4. RESULTS AND DISCUSSION

A case study demonstrating the methodology was conducted along a 240-mile segment of the Mississippi River, stretching from Baton Rouge to the Gulf of Mexico, which represents the busiest corridor of waterborne commercial activity in the United States. Data was collected from January to April 2024 (**Figure 4**). Data acquisition is discussed followed by model development and results.

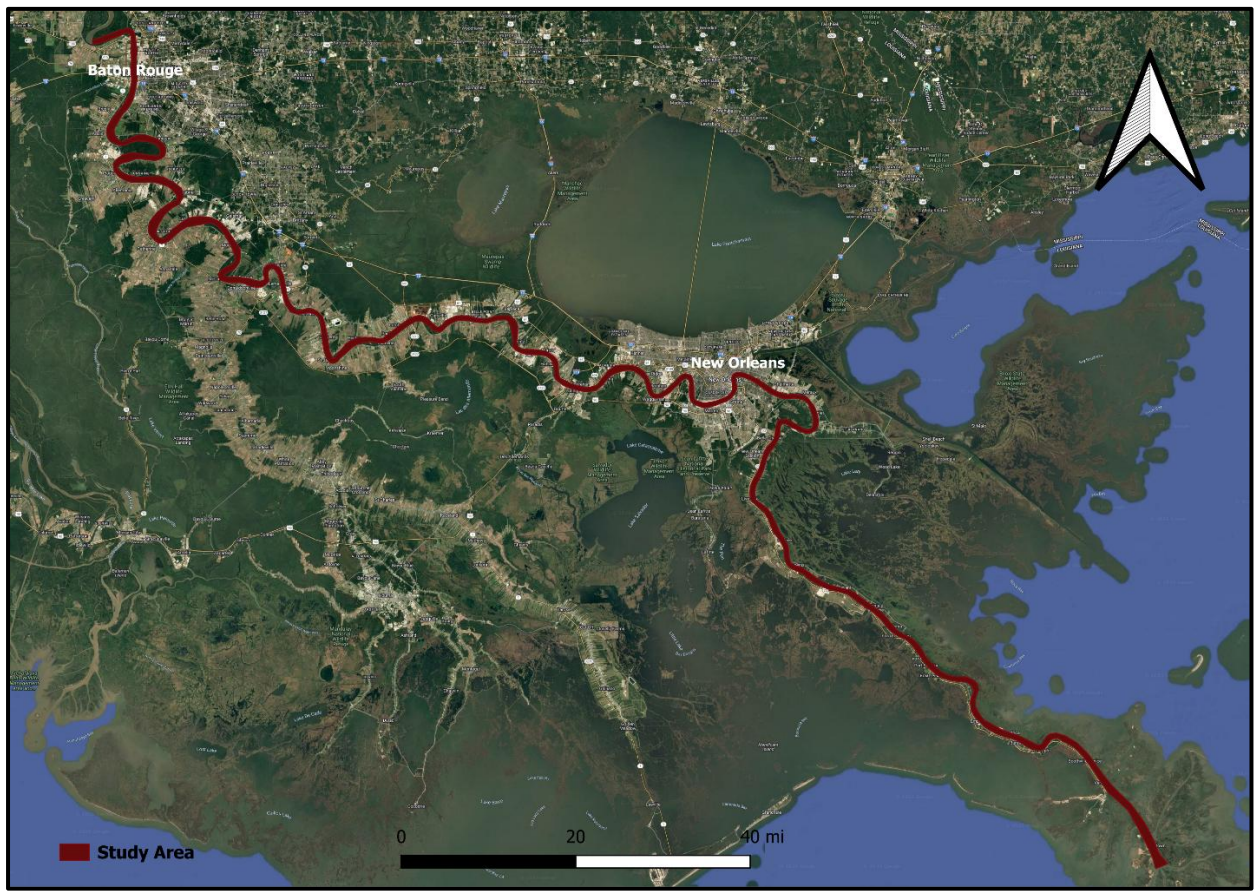

**Figure 4: Satellite map showing the study area along the Lower Mississippi River**

## 4.1. Data Acquisition and Image Annotation Results

Satellite imagery from Planet Labs (*35*) was used. It offers high resolution (3 m), frequently updated (daily with a 30-day delay) visual data, enabling detailed monitoring of inland waterways. To streamline data collection, we developed a custom Application Program Interface (API) that accepts AIS tracks for target regions, identifies spatial–temporal intersections with Planet imagery tiles, and automatically downloads only those scenes containing barges or vessels. These targeted downloads provided the labeled samples used to train our computer vision models. For this study, we used terrestrial AIS data sampled at 1-minute intervals across U.S. inland waterways obtained from the Marine Cadastre Web Tool (*36*). In relation to "dark vessels", although AIS is used to ensure vessel presence, the satellite tiles may still capture nearby "dark vessels" not broadcasting AIS.

A total of 688 images comprising approximately 4,550 labeled vessel and barge instances, covering an area of about 5,973 mi², was manually annotated from Planet satellite imagery using the Computer Vision Annotation Tool (CVAT) (*37*). These annotations included vessel types such as tugs/tows, cargo ships, and barge types (e.g., covered hoppers, crane barges) (**Figure 5**). Each instance was tagged with vessel and barge type and count, cover type, operational status, and direction of motion.

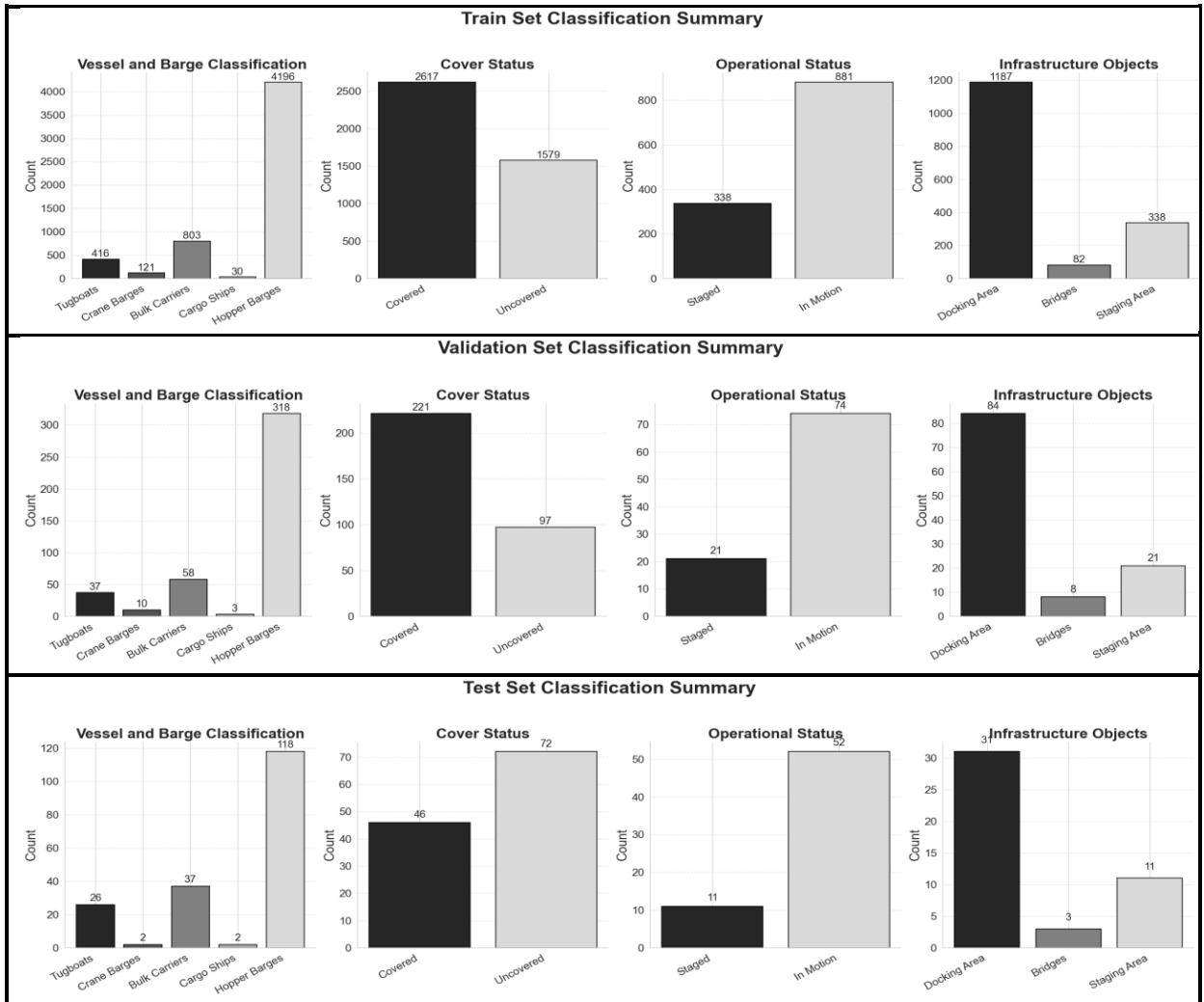

**Figure 5: Summary of Vessel, Barge and Infrastructure Type Distribution**

Data augmentation generated 698 additional images that were used for only model training. Data augmentation, such as horizontal flipping, random rotation (±10°), mosaic tiling, and color jittering, was applied to improve robustness to varying satellite image conditions, including lighting, angle, and background clutter (**Figure 6**). Horizontal flipping and rotation address variability in vessel orientation due to satellite viewing angles, mosaic tiling simulates complex spatial contexts by combining multiple scenes, and color jittering accounts for differences in lighting, atmospheric conditions, and sensor calibration.

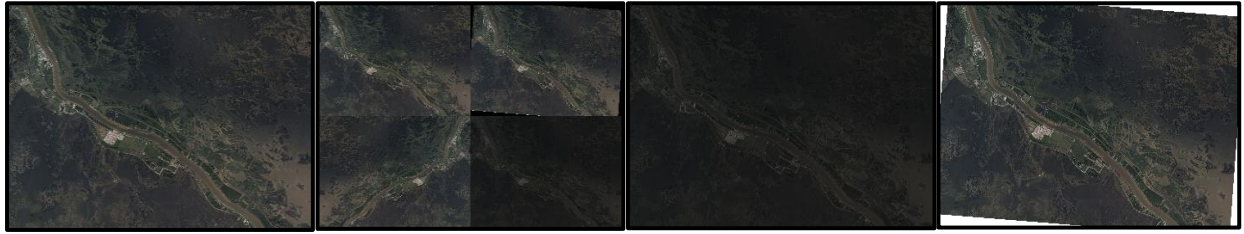

    a) Original      b) Mosaic      c) Jittering      d) rotation

**Figure 6: Data augmentation techniques employed**

### 4.2. Computer Vision Model Results

The YOLO model produces classifications for vessel and barge type, cover status, operational status, and direction of motion from satellite image alone (not using AIS). The dataset was randomly partitioned into training (70%), validation (20%), and testing (10%) subsets. The held-out test set, representing 10% of the data, comprises 176 vessel and barge instances.

A systematic random search was used to identify an effective combination of hyperparameters for the YOLOv11 model using the validation dataset. Learning rate, batch size, confidence threshold, and input image size were randomly sampled from predefined ranges and evaluated based on performance on a validation set. Training was conducted for up to 1000 epochs, with early stopping applied based on the validation loss trend (patience = 100 epochs). The AdamW optimizer with momentum was used, along with learning rate scheduling to adapt the update rate during training (**Table 1**).

**Table 1: Hyperparameters used for training YOLO models**

| Hyperparameter | Description | Tuning Range | Selected Value |
|---|---|---|---|
| **Learning Rate** | Controls speed of weight update iterations | 0.001 – 0.01 | 0.001 |
| **Batch size** | Number of samples processed per iteration | 2 – 32 | 8 |
| **Number of Epochs** | Complete dataset passes during training cycles | 100 – 1000 | 1000 |
| **Image Input Size** | Dimensions of input images fed network | 640 – 1280 | 800 |
| **IoU Threshold** | Minimum overlap ratio for positive detection | 0.4 – 0.6 | 0.5 |
| **Confidence Threshold** | Probability cutoff for accepting predicted detections | 0.4 – 0.6 | 0.5 |
| **Optimizer** | Algorithm adjusting network weights during training | Adam, SGD, AdamW, RMSProp | AdamW |

| **Early Stopping Patience** | Maximum epochs without improvement before stopping | 50 - 100 | 100 |
|---|---|---|---|

The model performs concurrent (non-hierarchical) classification of vessel and barge type, cover status, operational mode, and direction of motion directly from satellite imagery, without relying on AIS data. For example, a single object such as a barge can simultaneously be classified as a hopper barge, covered, in motion, and moving upstream. To facilitate performance assessment, the results are presented by detection category, with full details in **Table 2.** The discussion focused on F1-scores, as they represent a balanced measure of recall and precision.

**Table 2: Vessel Type, Cover Status, and Operational Mode Classification Results**

| Detection Category | Sub-class | No. of Instances | Precision | Recall | F1-Score |
|---|---|---|---|---|---|
| **Vessel and Barge Classification** | Tugboats | 17 | 0.992 | 1.000 | 0.996 |
| | Crane Barges | 2 | 0.933 | 1.000 | 0.965 |
| | Bulk Carrier | 37 | 0.959 | 0.973 | 0.965 |
| | Cargo Ships | 2 | 0.904 | 1.000 | 0.950 |
| | Hopper Barges | 118 | 0.990 | 0.852 | 0.915 |
| | **Overall** | **176** | **0.956** | **0.965** | **0.958** |
| **Cover Status** | Covered | 46 | 0.974 | 1.000 | 0.986 |
| | Uncovered | 72 | 1.000 | 0.758 | 0.863 |
| | **Overall** | **118** | **0.990** | **0.852** | **0.916** |
| **Operational Status** | Staged | 11 | 0.980 | 1.000 | 0.990 |
| | In Motion | 26 | 0.992 | 1.000 | 0.996 |
| | **Overall** | **37** | **0.988** | **1.000** | **0.994** |
| **Infrastructure Objects** | Docking Area | 31 | 0.974 | 1.000 | 0.987 |
| | Bridge | 3 | 0.994 | 1.000 | 0.997 |
| | Staging Area | 11 | 0.980 | 1.000 | 0.990 |
| | **Overall** | **45** | **0.983** | **1.000** | **0.991** |

*Vessel and Barge Classification*
Regarding *vessel and barge classification*, the overall F1-score of 95.8% across 176 vessel and barge instances. By vessel and barge type subcategories, tugboats had an F1-score of 99.6%, while crane barges and bulk carriers both showed F1-scores above 96.0%. Despite having fewer examples, cargo ships achieved an F1-score of 95.0%. Hopper barges, though the most prevalent in the dataset, had an F1-score of 95.1%, suggesting occasional under-detection, possibly due to occlusion or overlapping barge configurations.

*Cover Status Detection*
By *cover status*, the overall F1-score was 91.6% for 118 instances. While the model framework permits assigning both vessel type and a cover status to any detected object, cover classification is only applicable to specific barge types, primarily hopper barges which are typically covered or uncovered. This explains the discrepancy between 120 total barge instances and 118 instances where cover status was predicted. Covered barges achieved F1-score of 98.6%. Uncovered barges, while showing perfect precision, had a lower recall of 75.8%, and a resulting F1-score of 86.3% indicating some missed detections. This discrepancy may stem from visual ambiguity in partially covered or low-contrast conditions. Adding instances of uncovered barges could further enhance automated inventory tracking of barge types in satellite imagery (**Table 2**).

*Operational Status*
*Operational status classification*, distinguishing between "in motion", and "staged" vessels yielded the highest overall performance among the three categories across 37 instances. It is important to clarify that while the total number of annotated objects for barge type is 120, only 37 instances were used for operational status classification. For context, barge cover is a barge-specific attribute, and each barge in a tow may have a different cover status (e.g., some barges covered, others uncovered, **Figure 7**). Therefore, every barge is individually labeled with a cover status. In contrast, operational status is assigned at the tow level. A single label applies to the entire vessel group (tow) being pushed by a tugboat, as the whole tow is either "in motion", or "staged." Thus, each tow corresponds to one operational status label, regardless of how many barges are in the configuration (**Figure 7**). As a result, the total number of operational status labels is equal to the number of unique tows analyzed, which in this case is 37.

Both operational status sub-classes achieved perfect recall, and the F1-scores exceeded 99.0%. The model likely relied on contextual features such as wake patterns, proximity to infrastructure, and relative orientation to distinguish movement states. These results demonstrate that visual context from satellite imagery can be reliably used to infer vessel behavior.

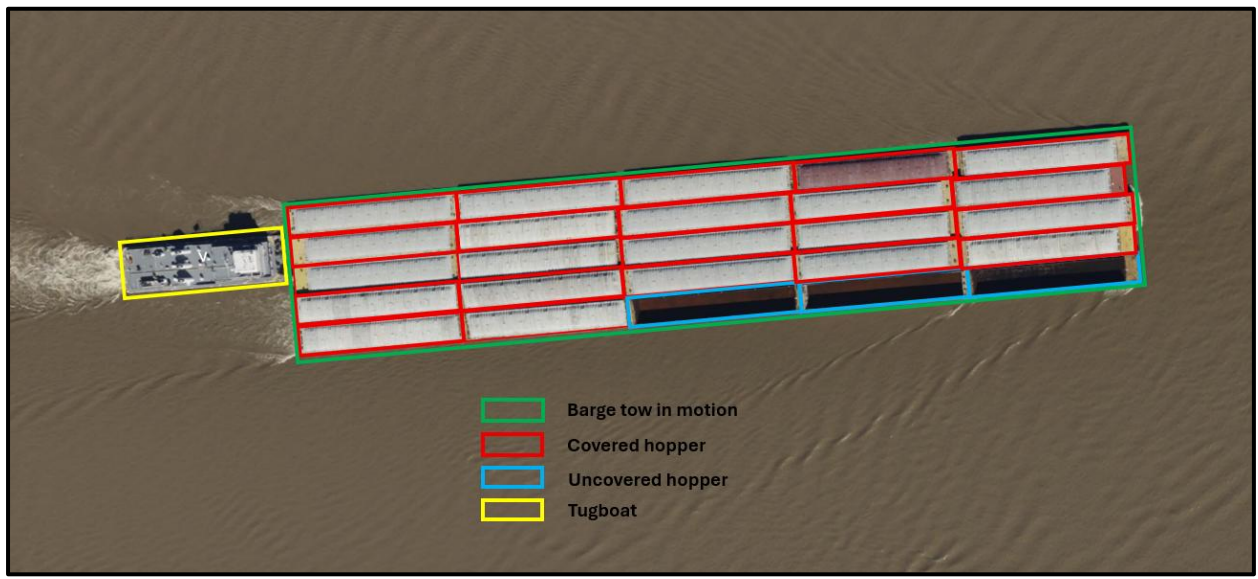

**Figure 7: Example of a single tow containing 25 barges that are covered and uncovered**

*Infrastructure Objects*
Infrastructure object detection achieved an overall F1-score of 99.1% across 45 instances (**Table 2**). Sub-class performance was consistently strong: docking areas achieved an F1-score of 98.7%, bridges reached 99.7%, and staging areas scored 99.0%. These high values reflect the distinct spatial and structural features of infrastructure elements, which are often easier to delineate in overhead satellite imagery compared to vessels or barges that may be partially occluded or visually similar. Notably, while vessels can vary in orientation and appearance, infrastructure objects exhibit consistent geometry and are typically fixed in location, contributing to robust detection.

*Barge Count Estimation*

The *barge count estimation* model achieved a sMAPE of 76% (**Figure 8**). On average, the predicted number of barges per tow was within a ±2 barge margin of error (e.g., MAE). The absolute error remains small in practical terms, especially given the complexity and variability of satellite imagery and vessel configurations. The model skews towards overestimation of barge count as shown by the scatter of points above the perfect prediction line. There is an evident outlier (38 barges) in the test data. Additional test cases involving large barge tows will be incorporated in the future to evaluate the model's performance across the full range of barge counts.

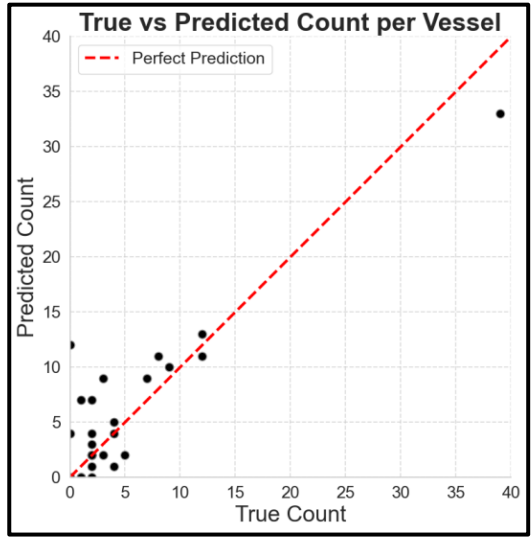

**Figure 8: Scatter plot comparing true and predicted vessel counts per observation**

*Direction of Movement Classification*

In the *direction of movement* classification, the model exhibited 100% accuracy in identifying upstream-moving vessels and 87.5% accuracy for downstream moving vessels (**Figure 9**). The discrepancy may be due to visual ambiguities in propulsion wake or vessel alignment in certain imagery scenes, particularly when vessels are moored near bends or in staging areas.

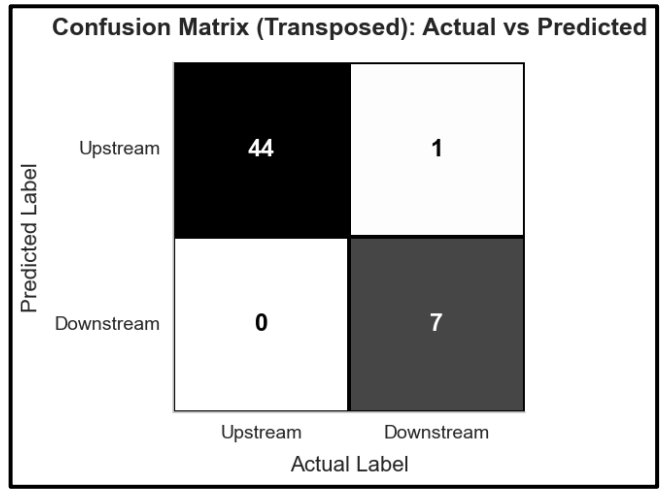

**Figure 9: Confusion matrix showing classification performance for vessel movement direction**

*Qualitative Assessment and Model Alternatives*

To qualitatively assess model performance, visual outputs from the trained YOLOv11 model are presented in **Figure 10**. Despite challenges such as occlusion, shadowing, and dense barge groupings, the model consistently produced precise bounding boxes and label assignments. These visualizations demonstrate the utility of the proposed framework for inland waterway monitoring and support its application in automated fleet tracking and infrastructure inventory generation.

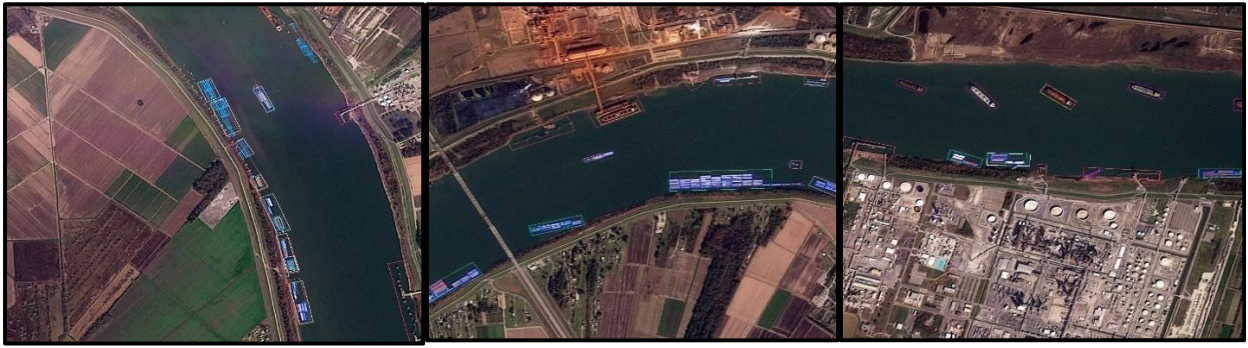

Figure 10: Sample Detections from Computer Vision Model

For comparison, the YOLOv8 model was also evaluated. YOLOv11 provided a marginal boost (~3–5%) in precision and recall across all detection categories, especially in challenging cases such as small or partially occluded barges. Key strengths included robust vessel detection in cluttered or turbid environments, accurate barge typing and status classification, and reliable count estimation for common configurations.

The largest sources of error were observed in **(Figure 11)** dense mooring areas, where overlapping barge boundaries resulted in merged detections or under-counts; and for staged barges, where operational status was harder to discern due to the absence of visible cues. These limitations suggest further gains could be achieved by incorporating temporal sequence models (e.g., object tracking) or leveraging AIS metadata during training (e.g., semi-supervised or multi-modal learning). Nonetheless, the framework can provide near-real-time situational awareness for riverine monitoring and transport planning tasks.

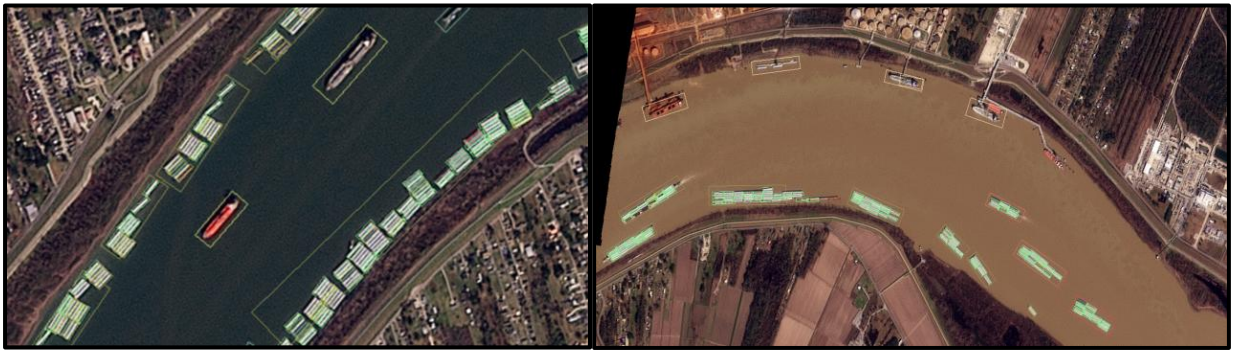

Figure 11: Examples of common sources of detection error

## 4.3. Spatiotemporal Fusion of Satellite and AIS Data Results

The results of linking satellite-detected vessels to AIS tracks across six imagery scenes (**Table 3**) show that, for each image, every satellite detection was matched successfully to its correct AIS record, yielding a linkage accuracy of 100 % in all cases. This perfect correspondence confirms that every vessel identified in the satellite imagery with active AIS transmissions was successfully associated with its AIS track. Because all satellite-detected vessels had corresponding AIS matches, no potentially dark vessels were identified in this study.

**Table 3: Spatiotemporal linkage results between satellite detections and AIS vessel tracks across six imagery scenes**

| Image ID | # Satellite Detected Vessels | # Linked Pairs | Linkage Accuracy (%) |
|---|---|---|---|
| 1 | 62 | 62 | 100 |
| 2 | 41 | 41 | 100 |
| 3 | 58 | 58 | 100 |
| 4 | 42 | 42 | 100 |
| 5 | 37 | 37 | 100 |
| 6 | 39 | 39 | 100 |

## 5. SPATIAL TRANSFERABILITY ANALYSIS

To assess the generalizability of the trained models, evaluations are conducted on satellite imagery from geographically distinct inland waterway systems. To evaluate the generalization capacity of the trained detection models beyond their original geographic domain, the models, trained on imagery from the Lower Mississippi River, were tested on inland waterway segments that featured different vessel traffic patterns, infrastructure layouts, and environmental characteristics. Performance degradation under domain shift indicates areas where domain adaptation or fine-tuning may be required to scale the framework across operational environments. The Atchafalaya River (Simmesport, LA) (**Figure 12**) was selected for spatial transferability. This river segment is a swamp-forested distributary of the Mississippi River with narrower channels and vegetated banks, providing a distinct riverbank structure and flow environment. Planet Labs imagery for the test region was retrieved using the same AIS-guided tile selection API employed during training data generation. This ensured consistency in vessel presence and annotation protocols. The models were evaluated on the same subtasks (e.g., vessel and barge classification, etc.) as in the original test set but without retraining the models on the test region.

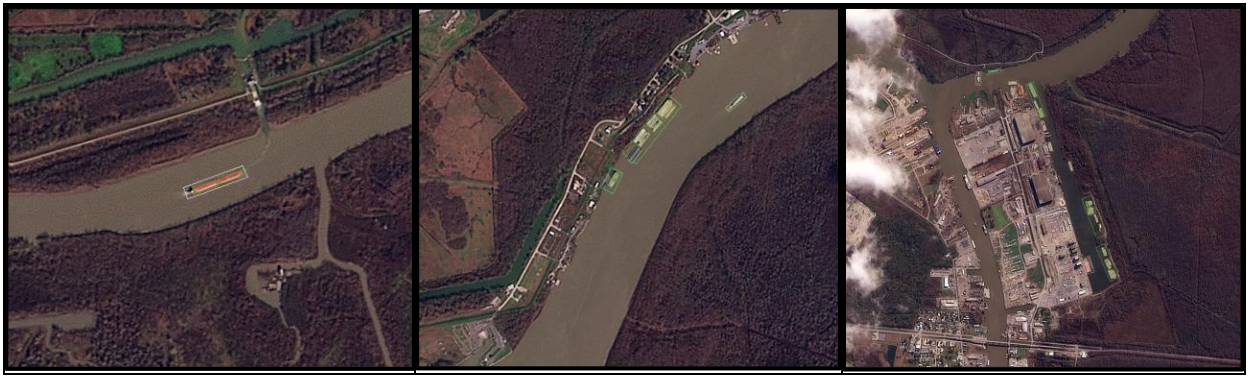

**Figure 12: Examples of river geography and topography along with vessel detections for test region (Atchafalaya River)**

Model performance decreased modestly (typically by 2–5%) (**Figure 12, Table 4**). The direction of movement showed 98% accuracy for the upstream accuracy and 94% for the downstream accuracy. Additionally, barge count estimation showed MAE of 2.8 barges and sMAPE of 80% for the test location.

The Object-to-AIS linkage showed 100% linkage accuracy with zero false positives and zero missing valid linkages.

**Table 4: Spatial Transferability of Computer Vision Model**

| Detection Category | Sub-class | Precision | Recall | F1-Score |
|---|---|---|---|---|
| **Vessel and Barge Classification** | Tugboats | 0.980 | 0.950 | 0.965 |
| | Crane Barges | 0.900 | 1.000 | 0.947 |
| | Bulk Carrier | 0.920 | 0.900 | 0.910 |
| | Cargo Ships | 0.850 | 1.000 | 0.919 |
| | Hopper Barges | 0.970 | 0.825 | 0.892 |
| | **Overall** | **0.944** | **0.935** | **0.939** |
| **Cover Status** | Covered | 0.950 | 0.967 | 0.958 |
| | Uncovered | 0.980 | 0.740 | 0.844 |
| | **Overall** | **0.970** | **0.825** | **0.892** |
| **Operational Status** | Staged | 0.900 | 0.900 | 0.900 |
| | In Motion | 0.950 | 0.933 | 0.941 |
| | **Overall** | **0.935** | **0.925** | **0.930** |

# 6. CONCLUSION

This study presents a comprehensive framework for enhancing inland waterway MDA by combining high-resolution Planet satellite imagery with state-of-the-art YOLOv11 object detection models. Through the creation of a manually annotated dataset of approximately 4,550 vessel and barge instances spanning approximately 5,973 mi², we demonstrated reliable detection and fine-grained classification of vessels and barges, including barge cover status, operational states (staged, in motion), barge count estimation, and upstream/downstream movement inference. The integration of transfer learning, systematic hyperparameter optimization, and rigorous model training and validation yielded mean Precision scores above 95% for vessel classification tasks, barge cover accuracy exceeding 95%, and direction-of-movement classification accuracy of 93%. Moreover, fusing these detections with AIS data achieved a 100% linkage accuracy, highlighting the potential for non-cooperative surveillance to fill critical AIS blind spots.

A spatial transferability analysis, testing the models on geographically disjoint river segments, revealed only a modest performance drop (2–5%), confirming that our approach generalizes well across diverse inland environments. These findings underscore the viability of deploying such a system for near-real-time monitoring of inland waterways, dynamic fleet inventories, and anomaly detection (e.g., dark vessels) in both research and operational settings.

Despite these promising results, several challenges remain. First, the training dataset, while carefully curated, remains relatively small and manually annotated; further scaling and semi-automated labeling (e.g., through AIS-guided pseudo-labels) could improve robustness. Second, our binary upstream/downstream movement classification does not capture more granular navigational behaviors (e.g., turning, station-keeping) and relies on sufficient contextual clues.

Future work will focus on expanding the annotated dataset to cover additional waterways and environmental conditions, integrating temporal sequence models for multi-object tracking, and exploring multi-modal deep learning architectures that jointly learn from imagery and AIS time-series. By addressing these limitations, we aim to develop an operationally scalable, end-to-end inland waterway surveillance system capable of delivering dynamic, high-precision maritime intelligence.

## 7. ACKNOWLEDGEMENTS

The authors acknowledge the support and sponsorship provided by the National Science Foundation (NSF), and the U.S. Army Corps of Engineers (USACE).

The authors acknowledge the use of OpenAI's ChatGPT to assist with language refinement. All analyses, results, and interpretations were conducted independently by the authors.

## 8. AUTHOR CONTRIBUTIONS

The authors confirm their contributions to the paper as follows: study conception and design: G. Agorku, S. Hernandez; data collection: G. Agorku, H. Hames and C. Wagner; analysis and interpretation of results: G. Agorku, S. Hernandez; draft manuscript preparation: G. Agorku and S. Hernandez.

## 9. DECLARATION OF CONFLICTING INTERESTS

The authors declare that they have no known competing financial interests or personal relationships that could have appeared to influence the work reported in this paper.

## 10. FUNIDNG

The authors disclosed receipt of the following financial support for the research: This research was supported by the National Science Foundation (NSF), Directorate of Engineering, Division of Civil, Mechanical, and Manufacturing Innovation [Award number 2042870].